\documentclass[a4paper]{svproc}

\usepackage[T1]{fontenc}
\usepackage{graphicx}
\usepackage{amsmath}
\usepackage{amssymb}
\usepackage{booktabs}
\usepackage{url}
\usepackage{xspace}
\usepackage{capt-of}
\usepackage{placeins}
\usepackage{hyperref}

\providecommand{\doi}[1]{\href{https://doi.org/#1}{doi:#1}}
\newcommand{\amp}{AMP\xspace}
\makeatletter
\def\@fnsymbol#1{\ensuremath{\ifcase#1\or \dagger\or \star\or
   {\star\star}\or \ddagger\or \mathchar "278\or \mathchar "27B\or
   \|\or **\or \dagger\dagger\or \ddagger\ddagger \else\@ctrerr\fi}}
\makeatother
\begin{document}
\mainmatter

\title{Learning Roller-Skating Motions of Humanoid Robots Based on Adversarial Motion Priors}
\titlerunning{Humanoid Robot Roller-Skating with AMP}

\author{Yunkang Cheng\inst{1}\thanks{These authors contributed equally.} \and
Yutong Wu\inst{1}\footnotemark[1] \and
Menghan Li\inst{1} \and
Shihe Zhou\inst{1} \and
Mingguo Zhao\inst{1,2,3}\thanks{Corresponding author.}}
\authorrunning{Y. Cheng et al.}
\tocauthor{Yunkang Cheng, Yutong Wu, Menghan Li, Shihe Zhou, and Mingguo Zhao}

\institute{
Department of Automation, Tsinghua University, Beijing, China
\and
Beijing Key Laboratory of Embodied Intelligence Systems, Beijing, China
\and
Institute for Embodied Intelligence and Robotics, Tsinghua University, Beijing, China\\
Corresponding author: Mingguo Zhao\\
Project website: \url{https://cyk579.github.io/Roller-Skating/}
}

\maketitle

\begin{abstract}
Humanoid roller-skating is difficult because the robot must coordinate whole-body balance, rolling contacts, and velocity-dependent posture regulation. This paper presents an adversarial motion prior based reinforcement learning framework for two humanoid roller-skating gaits: Pump Glide skating and Push Glide skating. The two gait datasets are collected independently through motion capture and retargeted to the humanoid robot separately. The retargeted data are then smoothed and resampled into reference motion states for AMP training. The two gaits are learned by independent AMP training pipelines with separate reference datasets, separate policies, and independent reward architectures. Simulation experiments are designed to evaluate gait quality, velocity tracking, turning, and gait-specific reward ablations.

\keywords{humanoid robot, roller-skating, reinforcement learning, adversarial motion prior}
\end{abstract}

\section{Introduction}

\begin{figure}[!t]
\centering
\begin{minipage}[t]{0.49\textwidth}
\centering
\includegraphics[width=\linewidth]{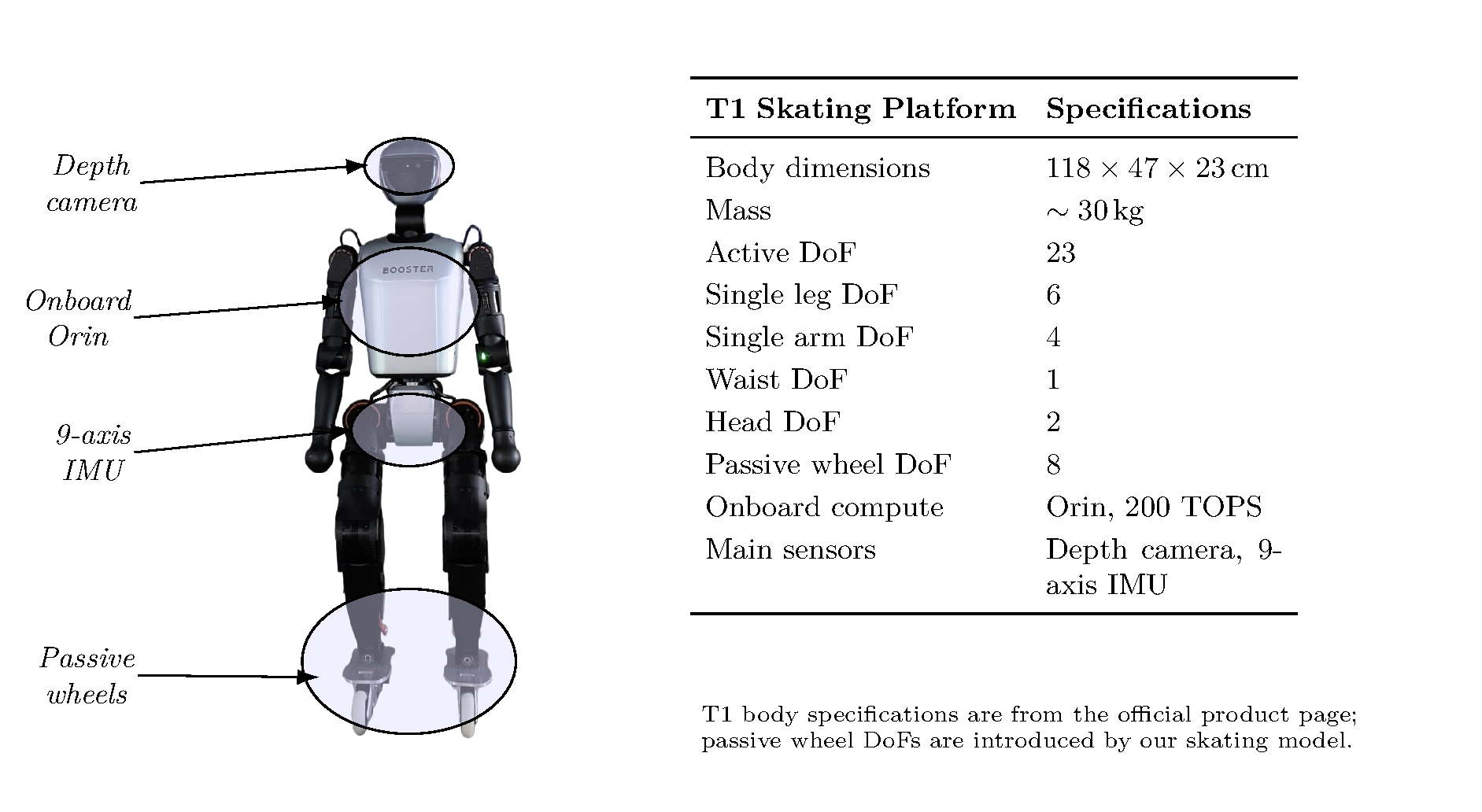}
\end{minipage}\hfill
\begin{minipage}[t]{0.49\textwidth}
\centering
\includegraphics[width=\linewidth]{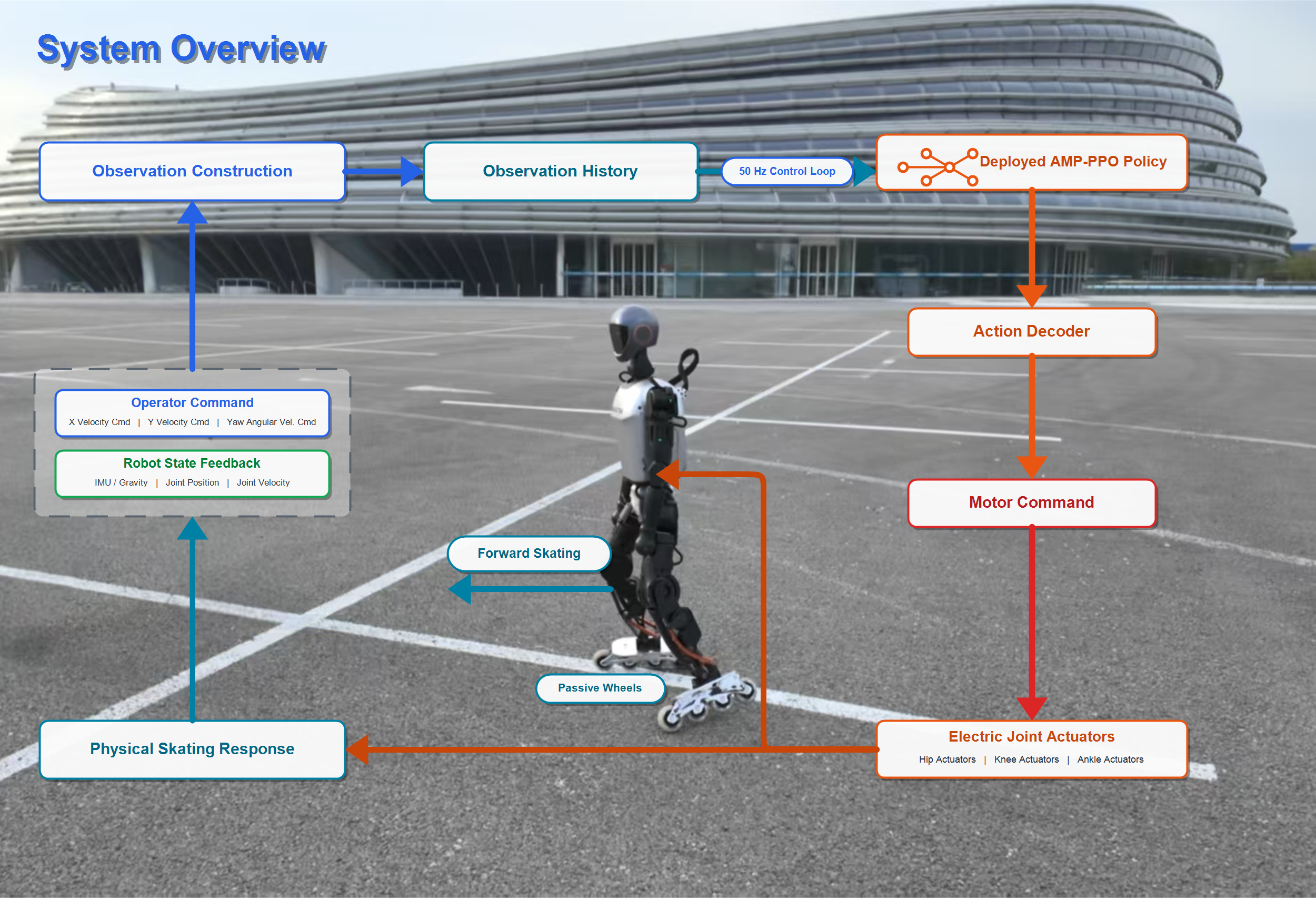}
\end{minipage}
\caption{T1 humanoid skating platform, passive wheel modification, and system overview.}
\label{fig:t1_system}
\label{fig:system_overview}
\end{figure}

This work is built around the experimental system shown in Fig.~\ref{fig:t1_system}: a Booster T1 humanoid retrofitted with passive roller skates and controlled by an \amp-PPO policy. The robot has 23 actuated body DoFs, while four unactuated skate wheels under each foot introduce 8 passive wheel DoFs. Operator velocity commands and robot state feedback are assembled into observation histories; the policy runs in a 50 Hz loop and outputs joint motor commands. Training and evaluation are conducted in Isaac Lab~\cite{mittal2025isaaclab} with the passive-wheel contact model described below.

Roller-skating provides a locomotion mode between bipedal walking and wheeled mobility: the robot preserves human-like leg motion while obtaining continuous gliding through skates. Wheeled-legged locomotion can be divided into actuated-wheel and passive-wheel systems; this paper focuses on the latter, where propulsion is generated by leg motion and wheel-ground friction rather than wheel motors. The goal is to realize humanoid roller-skating motions on this passive-wheel system, not only in simulation but also in real-robot trials.

This goal makes skate-wheel simulation a central technical challenge. Flat foot support is replaced by rolling contact, and the passive wheels introduce underactuated dynamics, anisotropic friction, velocity-dependent posture regulation, and gait-specific propulsion. Therefore, the fidelity and stability of the wheel model directly affect sim-to-real feasibility and evaluation accuracy. Although \amp has become a mature route for legged locomotion and character control, applying it to roller skating is still nontrivial because the learned style must remain compatible with rolling contact, weight transfer, and passive-wheel constraints.

Accordingly, this paper makes three contributions:
\begin{enumerate}
\item A passive skate-wheel simulation method for humanoid roller skating, including a sliced-cylinder collision model and geometric analysis of wheel-contact fidelity.
\item An \amp-based learning--simulation--deployment pipeline that converts human skating demonstrations into gait-specific motion priors and trains separate policies for Pump Glide and Push Glide.
\item Simulation and real-robot validation of two passive-wheel humanoid skating gaits, with analysis of velocity response, gait quality, and support timing.
\end{enumerate}

\section{Related Work}

\subsection{Humanoid Dynamic Motion Control}

Humanoid locomotion has long relied on model-based dynamics, trajectory optimization, and whole-body control. The linear inverted pendulum model and ZMP preview control provide low-dimensional stability descriptions for bipedal walking~\cite{kajita2003lipm}. On platforms such as Atlas, footstep planning, center-of-mass trajectories, state estimation, and whole-body optimization have been integrated for complex contact-rich tasks~\cite{kuindersma2016atlas}. Related ideas have also been extended to centroidal and whole-body quadratic-programming control~\cite{ficht2022direct}. These methods are interpretable and can explicitly handle balance and constraints, but in roller-skating the support point continuously moves with passive wheel rolling, while lateral push-off, wheel-ground friction, and body posture are strongly coupled. Accurate contact modeling and trajectory design therefore become expensive.

GPU-parallel simulation and reinforcement learning have recently reduced the cost of dynamic skill design. Large-scale PPO can train legged locomotion policies within minutes~\cite{rudin2022walk} and has been extended to humanoid velocity tracking~\cite{thibault2024velocity}, zero-shot sim-to-real humanoid learning~\cite{gu2024humanoidgym}, and control of human mobility devices~\cite{rajendran2024segway}. However, pure task rewards often require substantial reward engineering, curricula, and domain randomization; without motion priors, learned policies may look unnatural. Passive-wheel skating adds rolling degrees of freedom and low lateral friction to ordinary legged locomotion, motivating the use of demonstration-based style constraints.

\subsection{Wheeled-Legged and Skating Robots}

Wheeled-legged robots combine the efficiency of wheeled mobility with the terrain adaptability of legged systems, and can be broadly grouped into actuated-wheel and passive-wheel designs. Actuated wheels directly control wheel speed or torque; for example, ANYmal on Wheels uses ZMP, whole-body control, and motion planning for fast wheeled quadruped locomotion~\cite{bjelonic2019keeprollin}. Passive wheels or roller skates do not provide wheel-end actuation. Propulsion must arise from leg swing, body weight transfer, and wheel-ground friction, making the system more underactuated and more sensitive to contact constraints. A recent review also notes that passive wheeled-legged robots are mechanically efficient and simple, but harder to control than actuated-wheel systems~\cite{petri2026passive}.

Skating and skateboarding also require explicit motion style and phase structure. Human motion analysis reveals support switching, push-off, and glide phases in roller skating~\cite{chen2019roller}. Robot studies have used periodic rewards for humanoid skateboarding~\cite{thibault2024skateboarding} and task rewards with curricula for humanoid Pump Glide skating~\cite{gu2026skater}. Quadrupedal skating has further explored co-design of passive-wheel structures and control policies~\cite{wang2026qskater}. These studies demonstrate the potential of skating as efficient locomotion, but most focus on a single motion form or hand-crafted rewards. This paper studies Pump Glide and Push Glide, two humanoid skating gaits with distinct propulsion mechanisms, and uses motion priors to reduce reliance on manually specified phase rules.

\subsection{Imitation Learning and Adversarial Motion Priors}

Learning from demonstration reduces the exploration burden for complex motion skills. DeepMimic~\cite{peng2018deepmimic} learns physics-based character skills using frame-wise imitation and task rewards, but usually requires phase alignment to reference trajectories. Generative adversarial imitation learning~\cite{ho2016gail} learns expert distributions with a discriminator, and \amp further applies adversarial discriminators to unstructured motion clips~\cite{peng2021amp}. Thus, a policy can preserve demonstrated style without strictly tracking a single reference trajectory. This is important for humanoid robots because human demonstrations differ from robot morphology, joint limits, and contact conditions; direct frame-wise tracking can reduce stability.

Recent humanoid imitation work shows that large-scale human motion data can train natural whole-body control policies. ExBody decouples upper-body imitation from lower-body velocity tracking~\cite{cheng2024exbody}; HumanPlus builds a data pipeline from human shadowing to humanoid skills~\cite{fu2024humanplus}; and ALMI uses adversarial upper-lower-body policy learning to coordinate motion tracking and locomotion stability~\cite{shi2025almi}. In this work, a standard Gaussian Mixture Regression (GMR) pipeline is used to build skating reference motions~\cite{araujo2025gmr}, and PPO~\cite{schulman2017ppo} policies are trained in Isaac Lab~\cite{mittal2025isaaclab}. Task rewards constrain velocity, posture, and contact safety, while \amp rewards constrain the gait style of Pump Glide and Push Glide from human demonstrations.

\section{Humanoid Roller-Skating Task}

\subsection{Skate-Wheel Modeling and Simulation}

The roller skates must be modeled explicitly because this work uses passive free wheels: forward rolling is generated entirely by wheel-ground contact friction. The wheel joints provide no active drive torque. The wheel angular acceleration is mainly determined by the tangential contact force, i.e., $I\dot{\omega}=(r_c\times F_t)\cdot a$, where $r_c$ is the vector from wheel center to contact point, $F_t$ is the tangential contact force, and $a$ is the wheel-axis direction. The simulation model must therefore provide a stable effective rolling radius and contact moment arm, while avoiding lateral or roll-direction support that a real narrow wheel would not provide.

\begin{figure}[!htbp]
\centering
\includegraphics[width=0.76\textwidth]{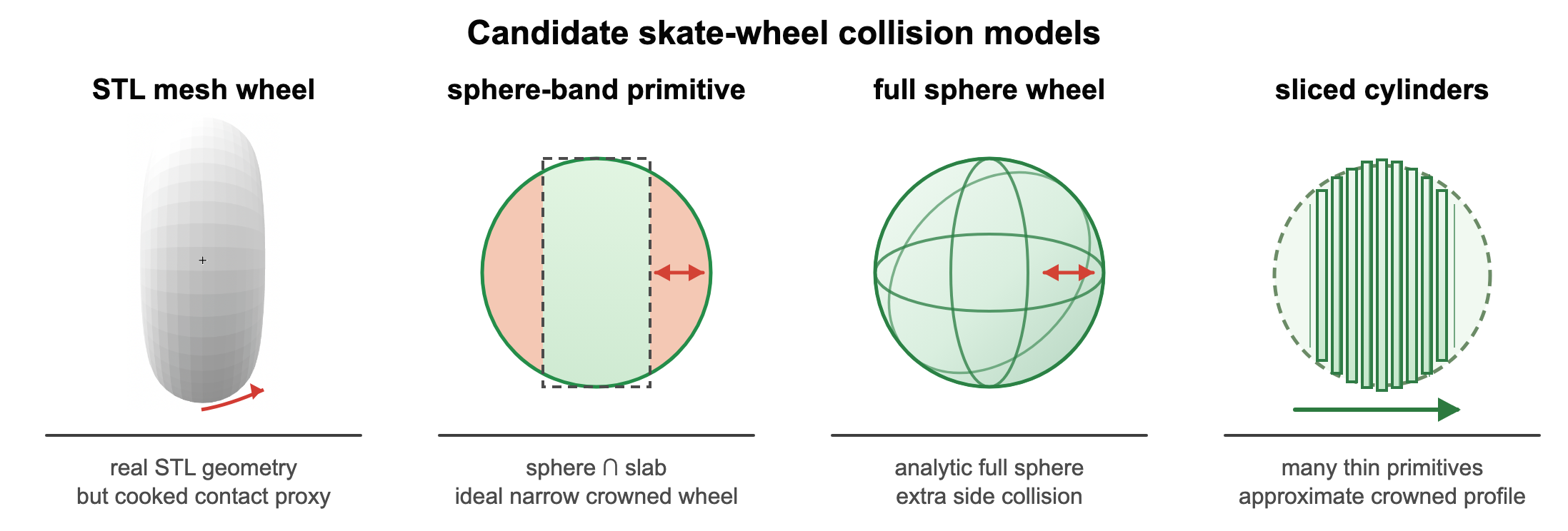}
\caption{Candidate collision models for skate wheels}
\label{fig:wheel_models}
\end{figure}

Figure~\ref{fig:wheel_models} compares four collision options. An STL wheel preserves visual geometry, but PhysX mesh cooking converts it into proxy collision geometry, and contact normals can jump, causing unstable passive rolling. A sphere-band primitive is closest to the actual rounded narrow wheel, but standard PhysX primitives do not directly support this CSG shape; converting it back to a mesh, convex decomposition, or SDF reintroduces the proxy-contact issue.

A full sphere provides analytic collision and stable rolling but has large geometric error. With wheel radius $R=32\,\mathrm{mm}$ and real wheel width $w=23.13\,\mathrm{mm}$, the sphere collision width along the wheel axis is $2R=64\,\mathrm{mm}$, about $2.77$ times the real wheel width. If a sphere-band is used as an upper-bound approximation of the true rounded wheel, the fraction of the full-sphere volume lying outside the true wheel width is
\begin{equation}
\eta_{\mathrm{extra}}
=1-\frac{\pi(R^2w-w^3/12)}{4\pi R^3/3}
\approx 48.2\%.
\end{equation}
Another error is the roll support point. The threshold at which the full-sphere lowest point leaves the real wheel-width range is approximately $\phi_{\mathrm{edge}}\approx21.2^\circ$, computed by $\phi_{\mathrm{edge}}=\sin^{-1}(w/2R)$. Beyond this roll angle, the sphere creates support outside the real wheel side surface, giving the policy unrealistic lateral support.

\begin{figure}[!t]
\centering
\includegraphics[width=0.72\textwidth]{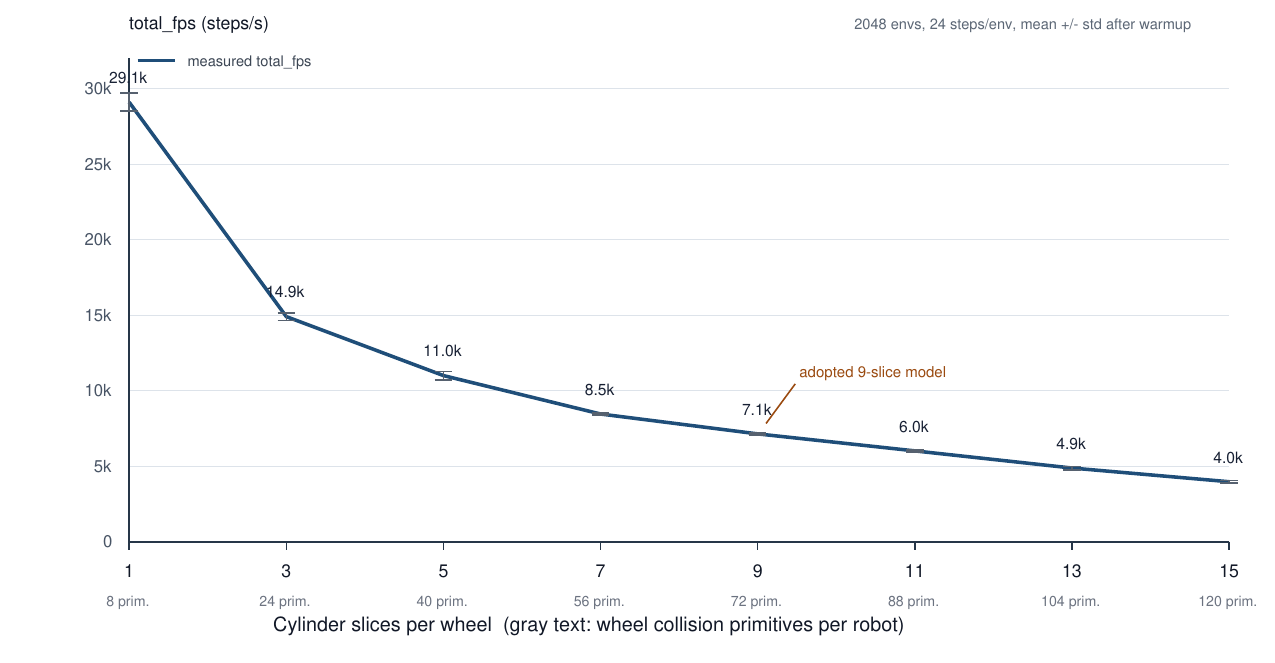}
\caption{Training throughput under different numbers of cylindrical slices}
\label{fig:wheel_slice_fps}
\end{figure}

We therefore use a sliced-cylinder wheel for training. The visual model remains the STL wheel, while the collision model is replaced by 9 narrow cylinders along the wheel width, with maximum radius $0.032\,\mathrm{m}$. This approximation provides stable rolling contact and avoids the large false support region of the full sphere, at the cost of more primitives and lower training throughput. Figure~\ref{fig:wheel_slice_fps} shows that 1, 3, 9, and 15 slices yield average throughputs of about $2.91\times10^4$, $1.49\times10^4$, $7.13\times10^3$, and $3.97\times10^3$ steps/s, respectively. We choose 9 slices as a compromise among rolling stability, geometric fidelity, and training cost.

\subsection{Gait Definitions and Task Objectives}

The control objective is to learn two policies, one for Pump Glide and one for Push Glide, so that the humanoid can execute the corresponding skating gait under a planar velocity command. At each control step, each policy receives its robot observation and task command and outputs the action required by the low-level joint controller, $a_t=\pi_{\theta}(o_t,c_t)$. The task command is $c_t=[v_x^{cmd},v_y^{cmd},\omega_z^{cmd}]^\top$, where the terms denote desired forward velocity, lateral velocity, and yaw rate.

Pump Glide is a periodic symmetric gait in which the two skates open and close to generate an hourglass-like trajectory. Its main challenge is to coordinate foot spacing, skate orientation, and body symmetry. Push Glide alternates between one leg pushing off and the other leg providing gliding support. Its main challenge is to coordinate support switching, push-off force, weight transfer, and gliding stability. In the experiments, both gait commands only use forward velocity; lateral velocity and yaw rate are set to zero.

\section{AMP-PPO Training for Two Skating Gaits}

The demonstrations of Pump Glide and Push Glide are obtained from motion capture. Since the human demonstrator and target robot differ in body proportions, joint limits, and skate-foot structure, we use a standard GMR pipeline to retarget human motion to robot reference motion~\cite{araujo2025gmr}. The trajectories are then smoothed, resampled, differentiated for velocity, and filtered to remove segments with ground penetration, self-collision, or abrupt joint motion. Passive wheel joints are not used as imitation targets. The processed continuous states are organized as short state-transition samples, forming separate \amp reference datasets for the two gaits.

\begin{figure}[!htbp]
\centering
\includegraphics[width=0.84\textwidth]{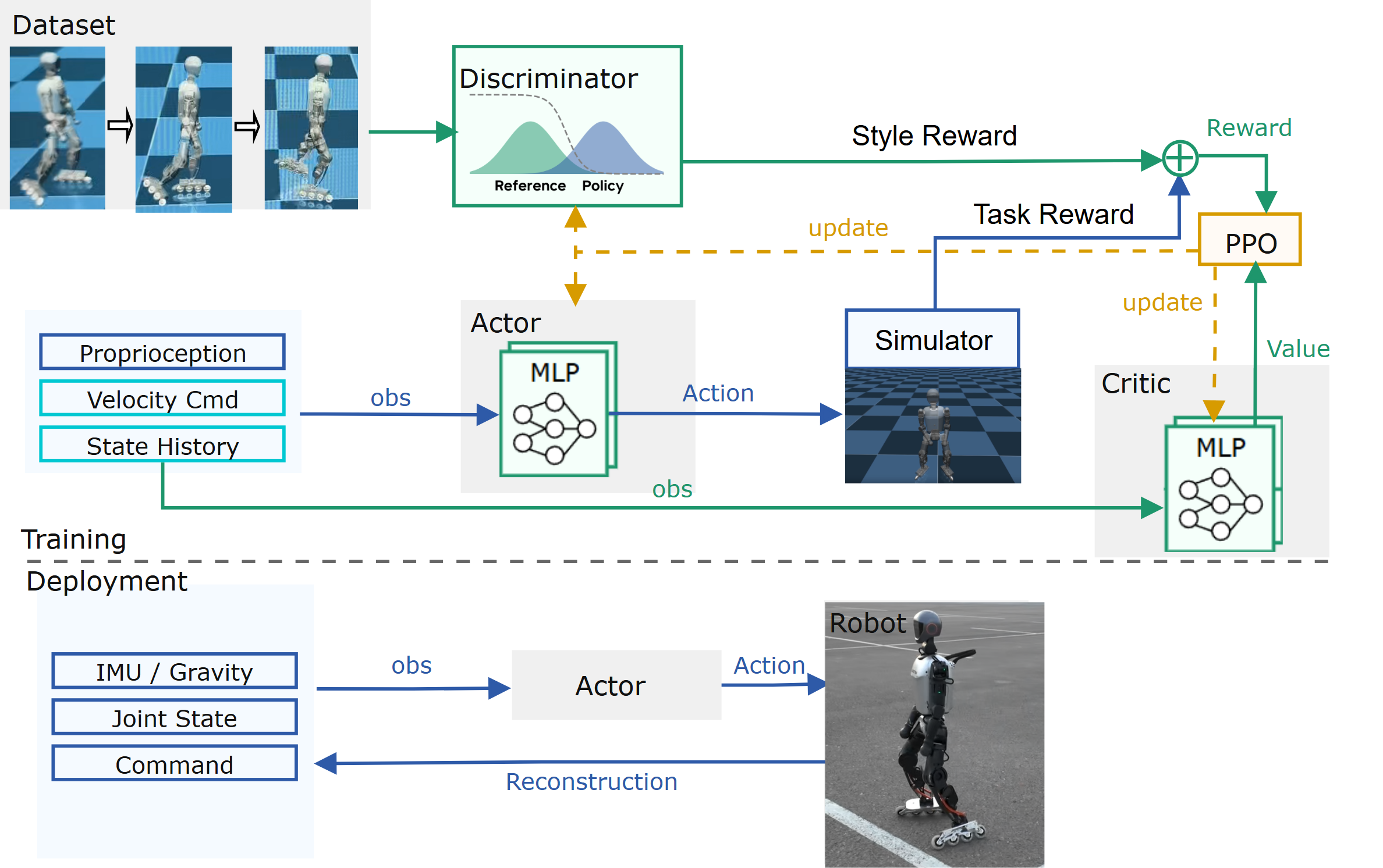}
\caption{\amp-based training framework for humanoid roller-skating gaits}
\label{fig:training}
\end{figure}

Based on these datasets, Pump Glide and Push Glide share the same AMP-PPO framework but use independent discriminators, policies, and task rewards because their propulsion and contact timing differ. Figure~\ref{fig:training} summarizes the training pipeline.

Both policies observe base angular velocity, projected gravity, velocity command, non-wheel joint states, and previous actions. The value network additionally observes root linear velocity and foot contact indicators. The policy outputs target positions for 21 non-wheel joints only; these targets are scaled by gait-specific action ranges and executed by a 50 Hz low-level PD controller. The wheel joints remain passive, so the policy focuses on balance, command response, and wheel-ground contact.

Command generation is gait-specific. Pump Glide uses a 12 s velocity curriculum covering zero-speed initialization, random forward skating in $[0.2,0.7]$ m/s, and stopping. Push Glide resamples commands every 12 s, with forward velocity in $[0,2]$ m/s, lateral velocity fixed to 0, and yaw rate in $[-0.2,0.2]$ rad/s.

For gait $m\in\{\mathrm{pump},\mathrm{push}\}$, the task reward is $r_{task,t}^{m}=\sum_i w_i^m\phi_i^m$. Velocity tracking uses first-order filtered velocities and exponential tracking terms:
\par\noindent\makebox[\textwidth][c]{\scriptsize
$v_t^f=0.9v_{t-1}^f+0.1v_t^{yaw},\quad
\omega_t^f=0.9\omega_{t-1}^f+0.1\omega_t^{world},\quad
\phi_{\exp}(e;\sigma)=\exp(-e^2/\sigma^2)$
}\par
Task rewards define measurable control objectives and safety boundaries. Pump Glide rewards mainly constrain velocity tracking, posture safety, foot-spacing bounds, and action regularization (left half of Table~\ref{tab:reward_weights}), without prescribing opening--closing phase. Push Glide additionally uses wheel air-time ratio, wheel speed, support-leg switching, and single-support duration terms (right half of Table~\ref{tab:reward_weights}) to provide contact and timing conditions. Its contact-timing terms use curriculum learning: the wheel air-time penalty decreases from 1.25 to 0.9 over 0--22000 simulation steps, suppressing early jumping strategies; the support-leg switching reward remains weak at 0.2 before 5000 steps and then linearly increases to 1.0 by 32000 steps, encouraging stable rolling before stronger alternating support. The motion style of both gaits is learned from the corresponding \amp discriminator.

\begin{table}[!htbp]
\centering
\caption{Task reward weights for Pump Glide and Push Glide}
\label{tab:reward_weights}
\begingroup
\scriptsize
\setlength{\tabcolsep}{3pt}
\renewcommand{\arraystretch}{0.86}
\begin{tabular}{@{}lr@{\hspace{1.8em}}lr@{}}
\toprule
\multicolumn{2}{c}{Pump Glide} & \multicolumn{2}{c}{Push Glide} \\
\cmidrule(r{1.2em}){1-2}\cmidrule(l){3-4}
Reward key & Weight & Reward key & Weight \\
\midrule
\texttt{survival} & 0.5 & \texttt{survival} & 0.5 \\
\texttt{track\_lin\_vel\_x\_exp} & 12.0 & \texttt{track\_lin\_vel\_x\_exp} & 12.0 \\
\texttt{track\_lin\_vel\_y\_exp} & 5.0 & \texttt{track\_lin\_vel\_y\_exp} & 3.0 \\
\texttt{track\_lin\_vel\_z\_l2} & -0.5 & \texttt{track\_lin\_vel\_z\_l2} & -0.5 \\
\texttt{track\_ang\_vel\_z\_exp} & 12.0 & \texttt{track\_ang\_vel\_z\_exp} & 2.0 \\
\texttt{ang\_vel\_xy\_l2} & -1.0 & \texttt{ang\_vel\_xy\_l2} & -3.0 \\
\texttt{energy} & $-2{\times}10^{-6}$ & \texttt{energy} & $-2{\times}10^{-5}$ \\
\texttt{dof\_vel\_l2} & -0.001 & \texttt{dof\_vel\_l2} & -0.0001 \\
\texttt{dof\_acc\_l2} & $-10^{-6}$ & \texttt{dof\_acc\_l2} & $-2{\times}10^{-7}$ \\
\texttt{action\_rate\_l2} & -0.01 & \texttt{action\_rate\_l2} & -0.004 \\
\texttt{undesired\_contacts} & -0.05 & \texttt{undesired\_contacts} & -0.05 \\
\texttt{flat\_orientation\_l2} & -0.7 & \texttt{flat\_orientation\_l2} & -0.5 \\
\texttt{feet\_too\_near} & -1.0 & \texttt{feet\_too\_near} & -0.02 \\
\texttt{feet\_too\_far} & -1.0 & \texttt{dof\_pos\_limits} & -0.6 \\
\texttt{dof\_pos\_limits} & -0.3 & \texttt{single\_stance\_time\_reward} & 0.05 \\
\texttt{body\_pitch\_penalty} & -0.7 & \texttt{wheel\_air\_time\_ratio} & -1.0 \\
 & & \texttt{support\_leg\_switch\_reward} & 1.2 \\
 & & \texttt{penalize\_single\_leg\_ahead} & -0.03 \\
 & & \texttt{body\_pitch\_penalty} & -0.35 \\
 & & \texttt{wheels\_spinning\_reward} & 1.0 \\
 & & \texttt{wheel\_drive\_torque\_penalty} & $-2{\times}10^{-4}$ \\
 & & \texttt{track\_lin\_vel\_y\_l2} & -0.5 \\
 & & \texttt{track\_ang\_vel\_z\_l2} & -0.6 \\
\bottomrule
\end{tabular}
\endgroup
\end{table}

\amp features are isomorphic to the reference motion and include body joint states, hand/foot key-point positions, and root velocities. Each discriminator receives a 5-frame concatenated \amp state, denoted by $z_t^m\in\mathbb{R}^{300}$.
The discriminator follows a Wasserstein form and outputs a scalar $D_\psi^m(z)$, with loss
\begin{equation}
\begin{split}
\mathcal{L}_{D}^{m} =
&-\mathbb{E}_{z\sim\mathcal{D}_{ref}^{m}}
\left[\tanh\left(\eta D_\psi^m(z)\right)\right]
+\mathbb{E}_{z\sim\mathcal{D}_{\pi}^{m}}
\left[\tanh\left(\eta D_\psi^m(z)\right)\right] \\
&+\lambda_{gp}\mathbb{E}_{\hat z}
\left[\left(\left\|\nabla_{\hat z}D_\psi^m(\hat z)\right\|_2-1\right)^2\right],
\end{split}
\end{equation}
where $\eta=0.4$, $\lambda_{gp}=10$, and $\hat z=\epsilon z^{ref,m}+(1-\epsilon)z^{\pi,m}$. The \amp reward is $r_{amp,t}^{m}=c_{amp}(1+\tanh(0.4D_\psi^m(z_t^m)))$, with $c_{amp}=2.0$. The PPO return combines task reward and style reward: for Pump Glide, $r_t^{pump}=0.40r_{amp,t}^{pump}+0.60r_{task,t}^{pump}$; for Push Glide, $r_t^{push}=0.45r_{amp,t}^{push}+0.55r_{task,t}^{push}$.

During training, 80\% of environments are initialized from random frames of the corresponding reference motion, while the rest start from the default pose. Observation noise, wheel-ground friction, mass, center of mass, PD gains, and wheel-joint damping are randomized. PPO uses an adaptive learning rate, target KL constraint, and gradient clipping.

\section{Gait Experiments}

Experiments combine simulation evaluations and real-robot trials on the modified T1 platform, reporting completion rate, travel distance, velocity error, torso tilt, foot-separation rhythm, support timing, and motion snapshots to evaluate sustained skating, command response, gait morphology, and support stability.

\subsection{Pump Glide}

\begin{figure}[!ht]
\centering
\includegraphics[width=0.62\textwidth]{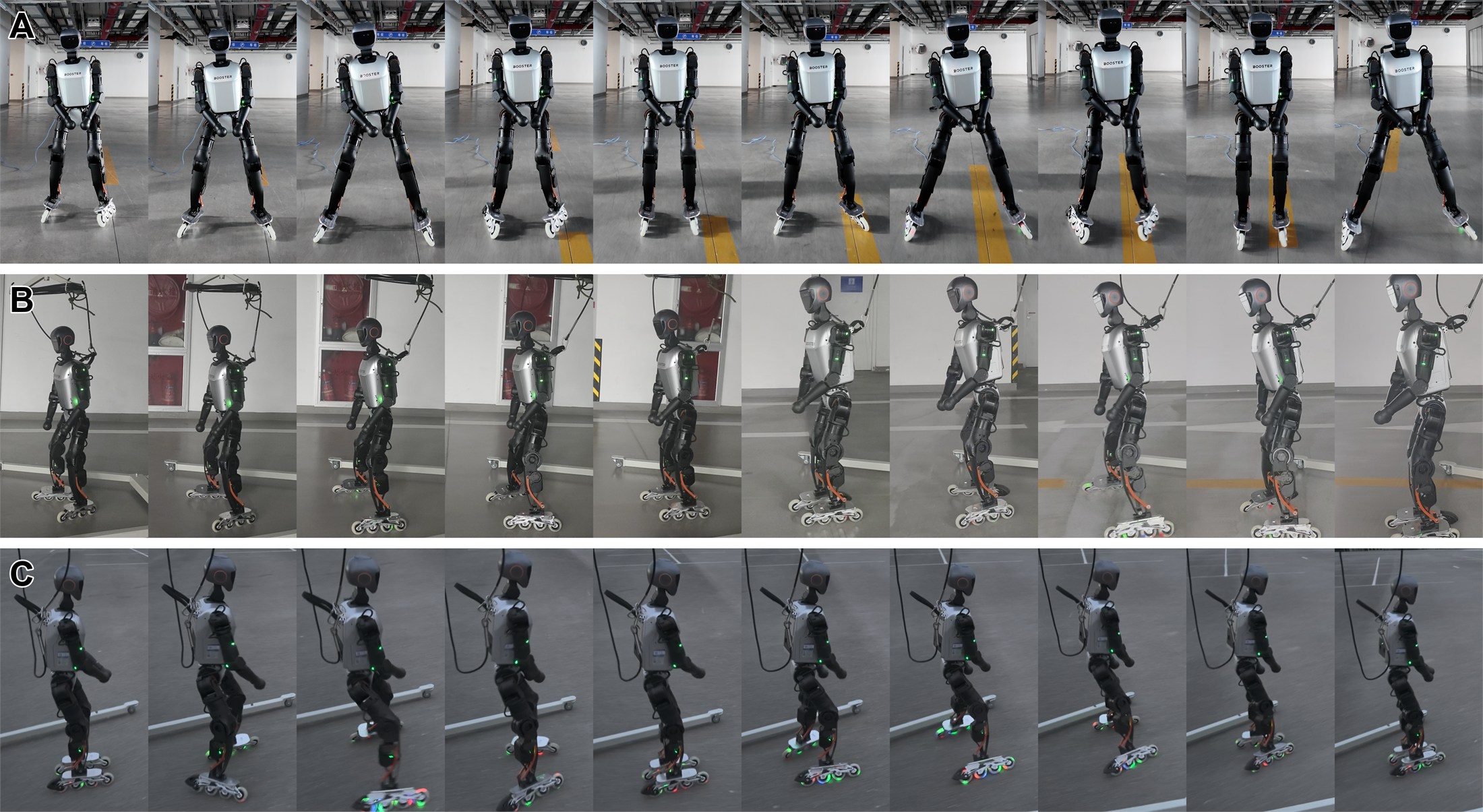}
\caption{Real-robot Pump Glide trials: low-friction front view (A), low-friction side view (B), and medium-friction side view (C)}
\label{fig:real_pump_glide}
\end{figure}

Figure~\ref{fig:real_pump_glide} shows that the robot completes opening, maximum spacing, closing, and minimum spacing phases within one cycle in real-robot trials. During forward skating, the upper body remains stable and the leg motion is symmetric. The rhythm mainly comes from the \amp style reward, while stability is bounded by task rewards.

\paragraph{Velocity command sweep.}
The robot first stabilizes at zero speed for 3 s and then tracks $v_x^{cmd}=0.10,0.15,\ldots,0.50\,\mathrm{m/s}$; each command is evaluated 64 times.

\begin{figure}[t]
\centering
\begin{minipage}[t]{0.49\textwidth}
\centering
\includegraphics[width=\linewidth]{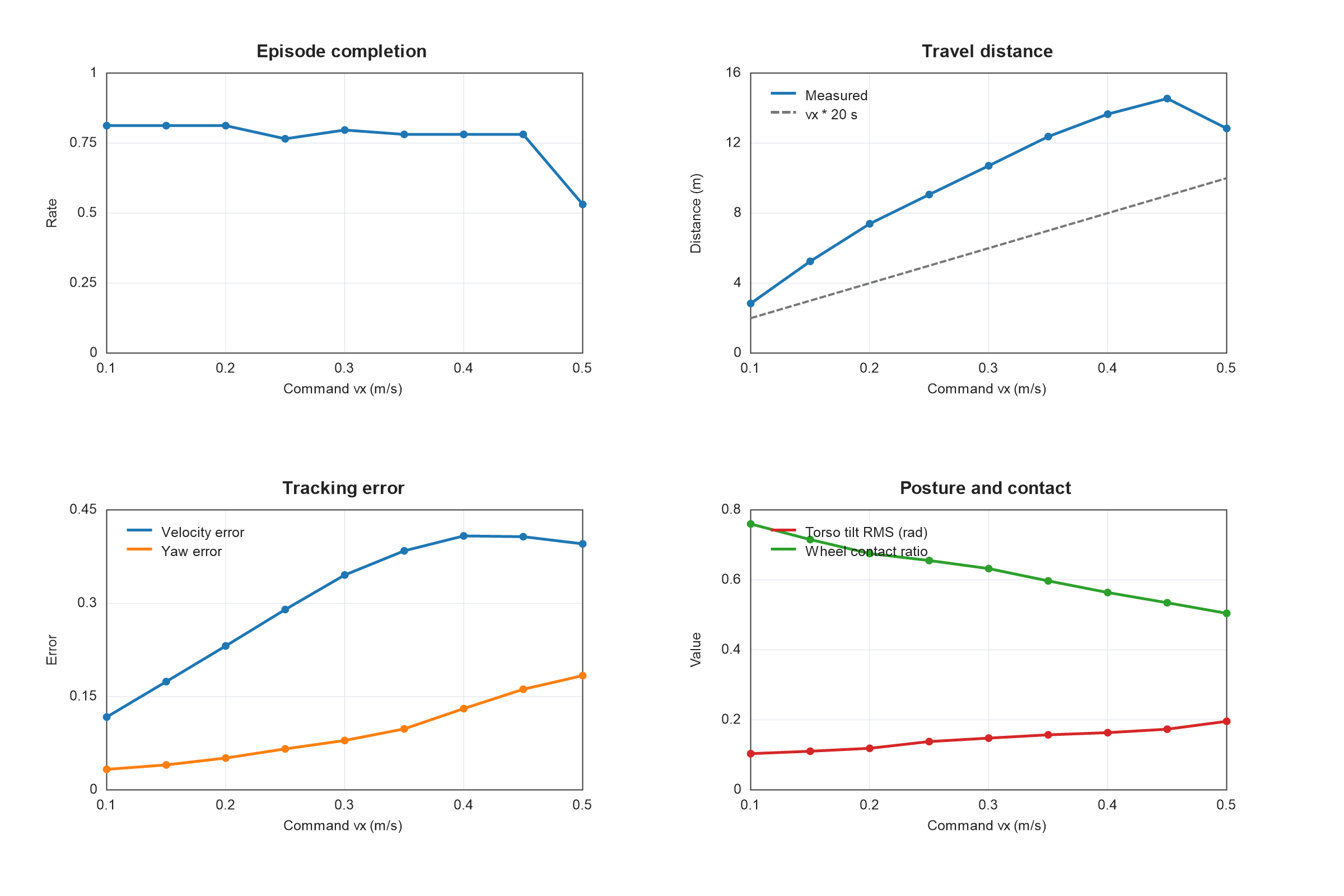}
\end{minipage}\hfill
\begin{minipage}[t]{0.39\textwidth}
\centering
\includegraphics[width=\linewidth]{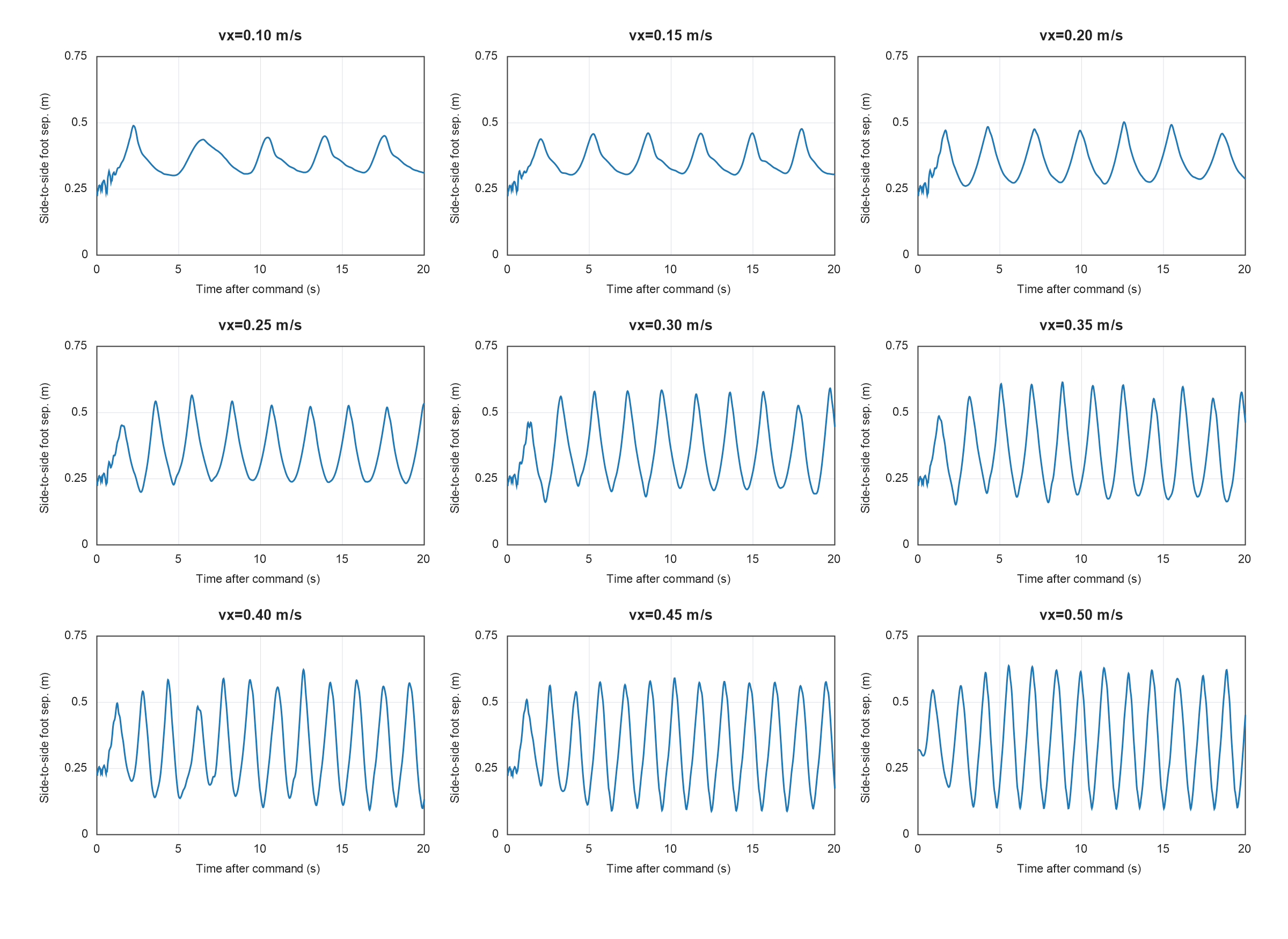}
\end{minipage}
\caption{Pump Glide results under different forward velocity commands. Left: velocity, completion, and posture metrics; right: lateral foot separation}
\label{fig:skate_x_speed_sliced_metrics}
\label{fig:skate_x_speed_sliced_foot_sep}
\end{figure}

Figure~\ref{fig:skate_x_speed_sliced_metrics} shows that commands of $0.10$--$0.45\,\mathrm{m/s}$ achieve completion rates of 0.766--0.813 and mean durations of 15.88--16.25 s; at $0.50\,\mathrm{m/s}$, the completion rate drops to 0.531. The maximum travel distance reaches 14.55 m, torso tilt RMS is 0.104--0.196 rad, and yaw-rate error is below 0.184 rad/s. Figure~\ref{fig:skate_x_speed_sliced_foot_sep} shows periodic modulation of foot separation with the velocity command, indicating that the policy adapts speed by adjusting the opening--closing rhythm.

\paragraph{Long-horizon velocity profile.}
The same policy tracks a 100 s profile that increases from 0.1 m/s to 0.4 m/s and then decreases to zero.

\begin{figure}[!ht]
\centering
\includegraphics[width=0.56\textwidth]{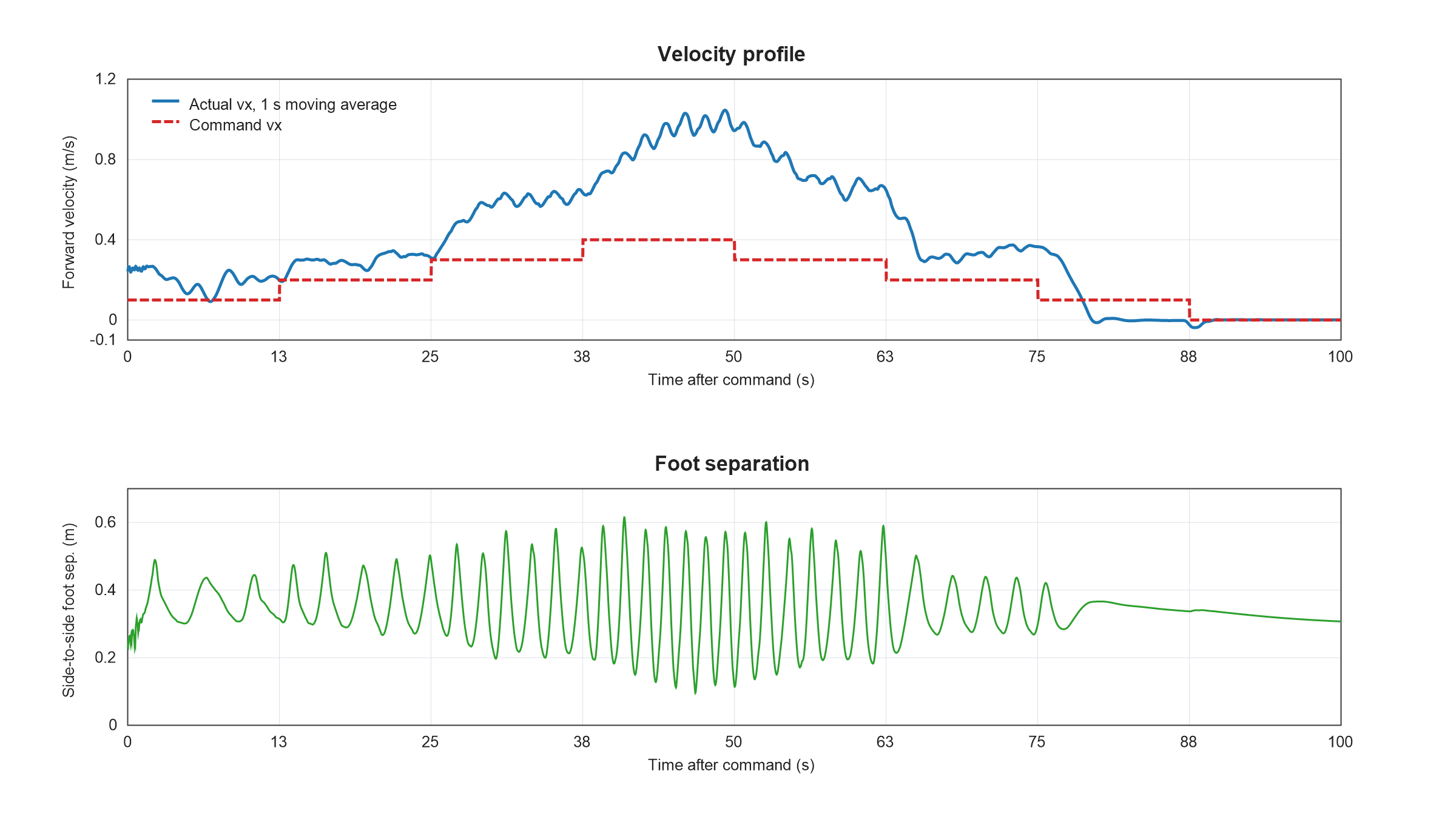}
\caption{Velocity response and lateral foot separation under a 100 s profile}
\label{fig:skate_x_profile_sliced_velocity}
\end{figure}

Figure~\ref{fig:skate_x_profile_sliced_velocity} shows a total travel distance of 39.50 m and torso tilt RMS of 0.132 rad; velocity follows the command trend while foot separation remains periodic.

\subsection{Push Glide}

\begin{figure}[!h]
\centering
\includegraphics[width=0.70\textwidth]{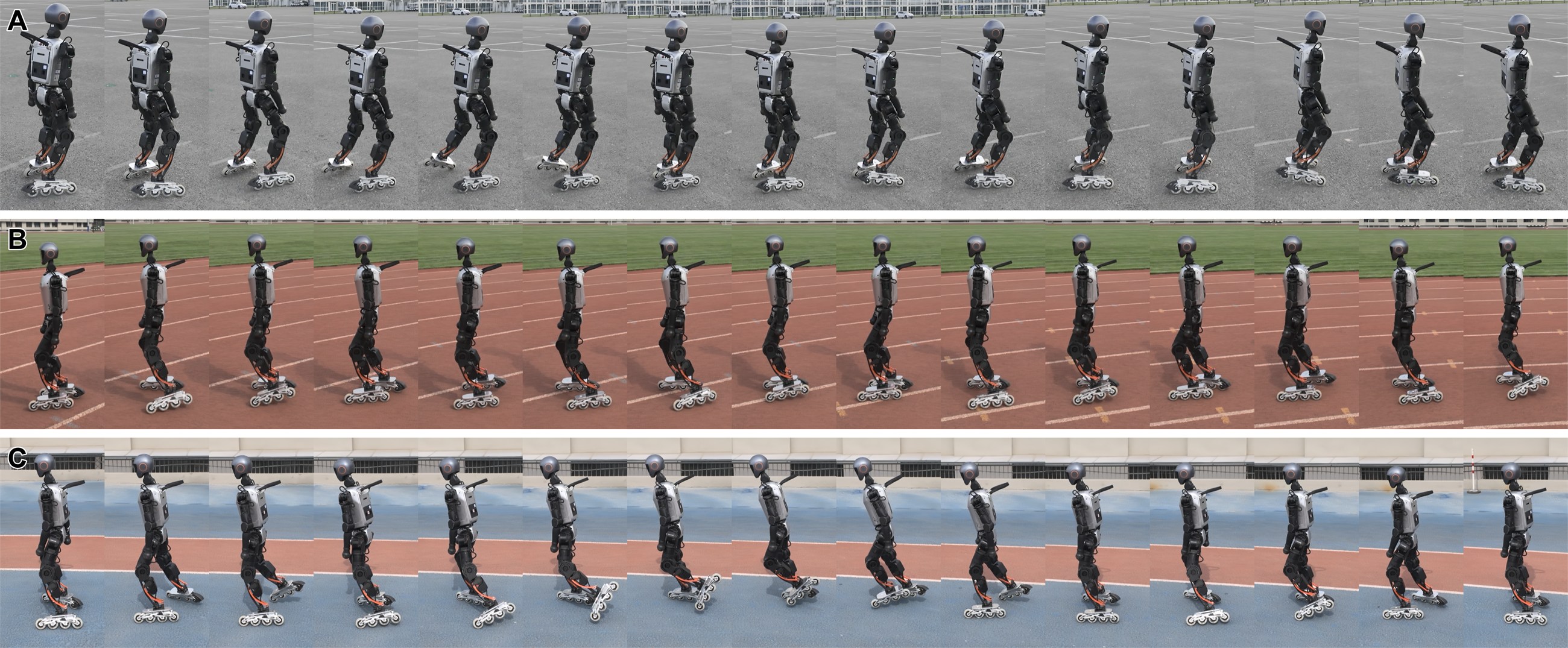}
\caption{Real-robot Push Glide trials}
\label{fig:real_push_glide}
\end{figure}

The Push Glide policy is tested in MuJoCo~\cite{todorov2012mujoco}. Each trial first stabilizes at zero speed for 3 s, then receives a target command; speed statistics use the stable window starting 8 s after command onset.
Figure~\ref{fig:real_push_glide} shows representative real-robot Push Glide motions. Unlike the symmetric opening-closing pattern of Pump Glide, Push Glide maintains a forward-leaning posture with fore-aft leg offset and alternating wheel-group support.

\paragraph{Forward velocity response.}
The forward velocity response only evaluates the velocity channel. Commands are set to $v_x^{cmd}=0.1$--$0.5\,\mathrm{m/s}$ with a step of $0.1\,\mathrm{m/s}$; lateral velocity and yaw-rate commands are zero. Velocity statistics use the first-order filtered forward speed in the yaw-aligned body frame, $v_{x,t}^{f}=0.9v_{x,t-1}^{f}+0.1v_{x,t}^{yaw}$. The velocity tracking error and yaw-rate error in the stable window are $e_v=T^{-1}\sum_{t=1}^{T}|v_{x,t}^{f}-v_x^{cmd}|$ and $e_\omega=T^{-1}\sum_{t=1}^{T}|\omega_{z,t}-\omega_{z,t}^{cmd}|$, respectively.

\begin{figure}[t]
\centering
\begin{minipage}[t]{0.54\textwidth}
\centering
\vspace{0pt}
\includegraphics[width=\linewidth]{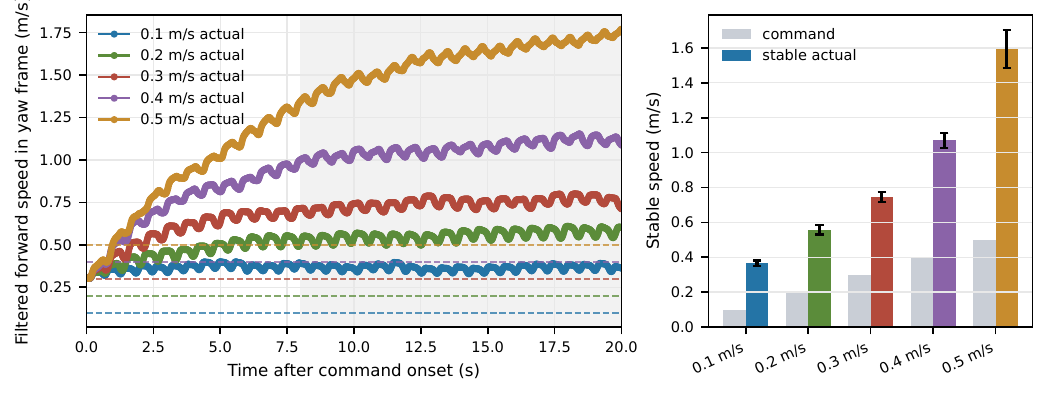}
\caption{Push Glide speed response under five commands}
\label{fig:skater_shuttle_speed_response}
\end{minipage}\hfill
\begin{minipage}[t]{0.36\textwidth}
\centering
\vspace{0pt}
\captionof{table}{Stable-window speed statistics}
\label{tab:skater_shuttle_speed}
\begingroup
\scriptsize
\setlength{\tabcolsep}{2pt}
\begin{tabular}{@{}cccc@{}}
\toprule
Cmd. & Actual $v_x^f$ & $e_v$ & Time \\
\midrule
(m/s) & (m/s) & (m/s) & (s) \\
\midrule
0.10 & $0.366 \pm 0.015$ & 0.266 & 20.00 \\
0.20 & $0.558 \pm 0.028$ & 0.358 & 20.00 \\
0.30 & $0.746 \pm 0.030$ & 0.446 & 20.00 \\
0.40 & $1.069 \pm 0.043$ & 0.669 & 20.00 \\
0.50 & $1.594 \pm 0.109$ & 1.094 & 20.00 \\
\bottomrule
\end{tabular}
\endgroup
\end{minipage}
\end{figure}

Figure~\ref{fig:skater_shuttle_speed_response} and Table~\ref{tab:skater_shuttle_speed} show that all five forward commands sustain 20 s simulations, and the actual speed increases monotonically with the command. However, the actual speed is consistently higher: from 0.10 to 0.50 m/s commands, actual speed increases from 0.366 to 1.594 m/s, and error increases from 0.266 to 1.094 m/s. The policy therefore achieves stable Push Glide and command sensitivity, but the speed channel still has a gain bias, possibly caused by passive-wheel contact coupling and Isaac/MuJoCo wheel-model differences.

\paragraph{Support-phase timing.}

\begin{figure}[t]
\centering
\includegraphics[width=0.76\textwidth]{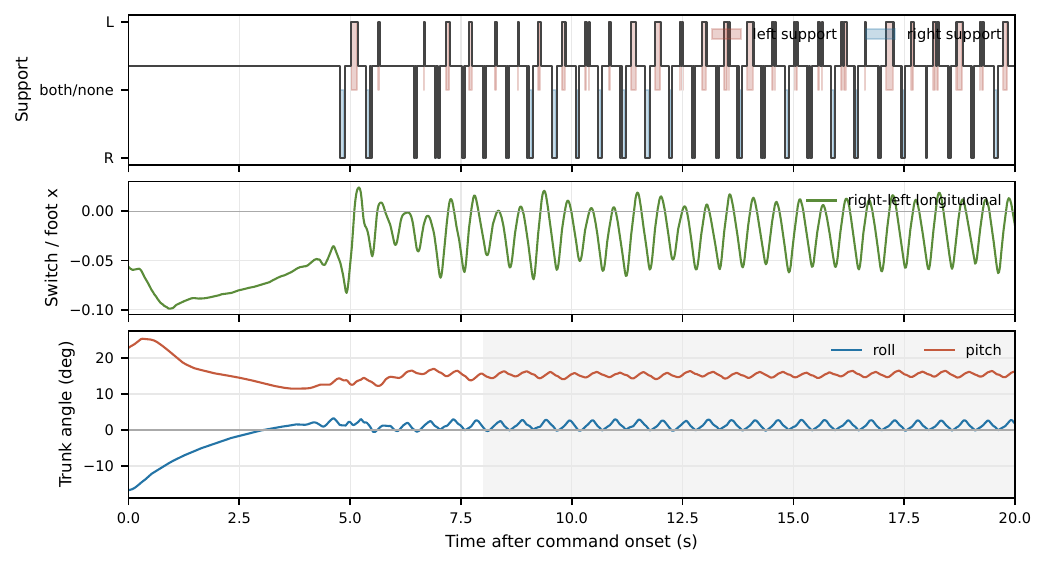}
\caption{Push Glide support phase and torso attitude}
\label{fig:skater_shuttle_support_phase}
\end{figure}

Figure~\ref{fig:skater_shuttle_support_phase} shows alternating short single-support intervals of left and right wheel groups, synchronized fore-aft foot phase oscillation, and bounded torso roll and pitch. This indicates that the support-leg switching reward encourages alternating support and rhythmic center-of-mass transfer instead of a fixed posture.

\section{Discussion and Conclusion}

This paper presents an \amp-PPO framework for passive-wheel humanoid roller skating, including a sliced-wheel simulation model, demonstration-based gait priors, and separate policies for Pump Glide and Push Glide. Experiments show sustained skating motions for both gaits: Pump Glide exhibits periodic foot opening--closing and stable long-horizon motion, while Push Glide achieves command-sensitive forward speed and alternating support timing, with representative real-robot trials for both gaits.

Limitations remain: the current evaluation is task-level rather than a standardized skating-quality benchmark, and the sliced-wheel model is still an approximation of real wheel-ground contact. Although its support-height discontinuity is small and its passive rolling stability is better than STL collision, mismatch among the approximation, deployment wheel models, and real contact may contribute to velocity-tracking errors in both Pump Glide and Push Glide. More accurate wheel-contact modeling, system identification, and closed-loop speed control remain future work.

\section*{Acknowledgements}

\begin{sloppypar}
\indent This research was supported by STI 2030-Major Projects 2021ZD0201402, Beijing Natural Science Foundation (L243004), National Innovation and Entrepreneurship Training Program for College Students (No. 95), Tsinghua University Undergraduate Academic Research Promotion Program, and Spark Program (``Science and Technology Innovation, Spark the Prairie Fire'' Tsinghua University Student Innovation Talent Training Program).
\end{sloppypar}

We thank Booster Robotics for providing the robot platform, experimental environment, and technical support. We thank SongRui Huang, Zhihan Li, Yiyi Zheng, Yihui Zhang, and Lushu Yang for assistance in data collection and experiments.

\end{document}